\documentclass[11pt]{article}
\usepackage[letterpaper,margin=1in]{geometry}
\usepackage[T1]{fontenc}
\usepackage[utf8]{inputenc}
\usepackage{times}
\usepackage{microtype}
\usepackage{amsmath,amssymb}
\usepackage{graphicx}
\usepackage{booktabs,multirow,array,tabularx,makecell}
\usepackage[table]{xcolor}
\usepackage{tikz}
\usetikzlibrary{arrows.meta,calc}
\usepackage{enumitem}
\usepackage{caption}
\usepackage{placeins}
\usepackage[round,sort&compress]{natbib}
\IfFileExists{xurl.sty}{\usepackage{xurl}}{\usepackage{url}}
\usepackage[colorlinks=true,linkcolor=blue!50!black,citecolor=blue!50!black,urlcolor=blue!50!black]{hyperref}
\usepackage[capitalise,nameinlink]{cleveref}
\crefname{figure}{Figure}{Figures}
\Crefname{figure}{Figure}{Figures}

\setlist{itemsep=2pt,topsep=4pt}
\newcommand{\gpt}{GPT-6-Astra}
\newcommand{\opus}{Opus~5}

\title{\textbf{RSI-Atlas: From Paper Chains to Verified Research Steps}\\[6pt]
\large Technical Report, Version 1}
\author{RSI AI4AI Research Team}
\date{November 2026}

\begin{document}
\maketitle

\begin{abstract}
Recursive self-improvement asks AI systems to make research progress that later rounds can build on. We call the unit of such progress a verified research step: an improvement over a specific prior method, made under a frozen protocol and confirmed by an evaluator that runs the submitted code itself. This report asks whether the published literature can supply tasks for such steps in quantity, whether current agents can take them, and what confirming a step requires. The approach rests on an observation: when a later paper on the same task and benchmark cites an earlier one and names a limitation of it, the literature has already taken a step, and the later paper can serve as a hidden reference for a task that starts from the earlier one. The Paper Atlas maps 7,317 papers and places 4,959 of them in 630 task--dataset clusters that contain 1,236 citation chains, and a task procedure turns a chain segment into a task with a frozen protocol and an evaluator that runs every submission. In a pilot of 20 tasks on six chains, Codex running GPT-6-Astra improved on the starting point in 12 tasks and Claude Code running Opus~5 in 10, plus 2 more after the operators repaired the submitted code. Which steps the agents took says more than how many. On tasks that ask for a cheaper method at fixed or bounded quality, 16 of 18 submissions improved, two of them after repairs. On tasks that ask for a better method, 8 of 18 comparable submissions did: 7 of 8 where the task's reference was more than 10\% better than the starting point, none of 6 where a later paper was less than 4\% better, and 1 of 4 where the reference value was the starting value. The submissions also separated metrics from claims. On the two chains where agents ran their own ablations, 3 of the 12 counted improvements came with an ablation that contradicted the proposed mechanism, and across the pilot 6 counted improvements took the same route as the task's reference. Finally, the size of a published step depended on the protocol under which it was measured. A verified step should therefore record, beside the metric confirmed by execution, whether the agent's own controls support its explanation and whether its route is new.
\end{abstract}

\section{Introduction}\label{sec:intro}

Recursive self-improvement, in the form we work toward, is a loop in which AI systems improve AI methods and each improvement becomes the starting point of the next round of research. Such a loop compounds what each round gets right and what it gets wrong, so it needs a unit of progress that later rounds can trust. We call this unit a verified research step: an improvement over a specific prior method, made under a frozen protocol and confirmed by an evaluator that runs the submitted code itself instead of reading the agent's report. This report asks three questions about such steps: whether the published literature can supply tasks for them in quantity, whether current agents can take them, and what confirming a step requires. This first version answers them for single steps; later versions chain the steps into a loop.

Current evaluations of research agents test parts of this unit. Reproduction benchmarks ask an agent to replicate a published result \citep{starace2025paperbench,siegel2024corebench,chen2025scienceagentbench}. Research-engineering benchmarks pose a competition or a starting solution with a metric to improve \citep{chan2025mlebench,huang2024mlagentbench,nathani2025mlgym,wijk2025rebench,zhang2025mlrcbench}. End-to-end systems write whole papers that human or automated reviewers then judge \citep{lu2024aiscientist,yamada2025aiscientistv2}, and evolutionary coding agents improve programs against automated evaluators on problems chosen by their authors \citep{novikov2025alphaevolve}. A verified step drawn from the literature needs three things at once: a starting method that the field has used, a limitation of that method stated in the literature, and an improvement that an evaluator confirms by running the agent's work. Each of these evaluations supplies only some of the three. Reproduction benchmarks start from the literature but ask for no improvement; research-engineering benchmarks and evolutionary agents confirm improvements by execution but start from competitions or problems their authors chose; end-to-end systems are judged by reviewers rather than by running the work. Supplying all three at scale means solving four problems: finding where the literature records an improvement, pinning down what was improved under which protocol, keeping the known answer out of the task, and running what the agent submits instead of trusting its report.

\begin{figure*}[t]
  \centering
  \input{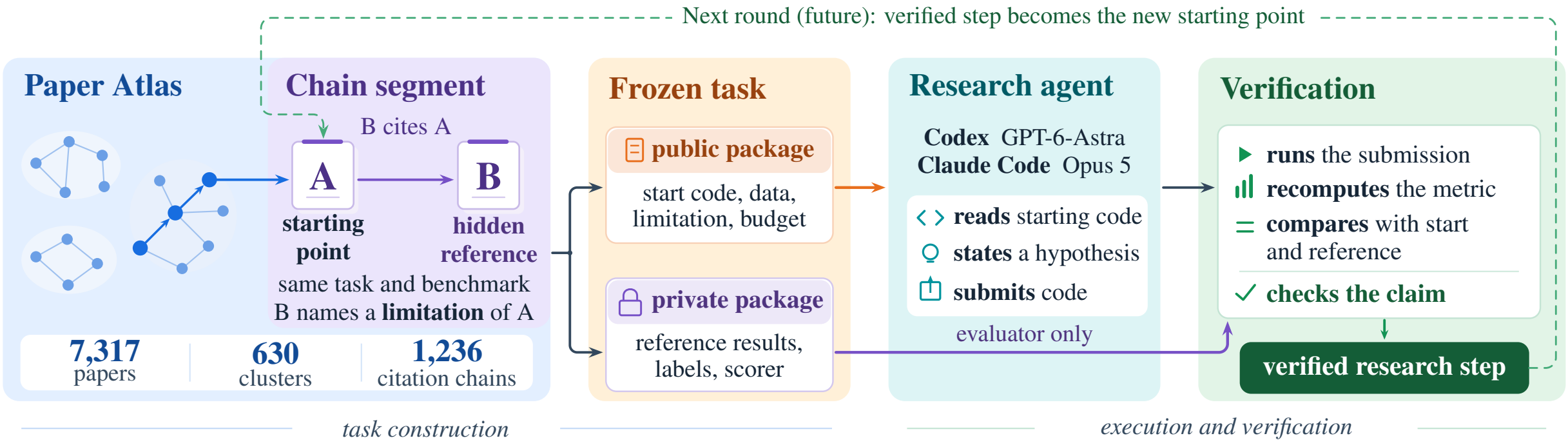}
  \caption{From the literature to a verified research step. A builder takes a chain segment from the Paper Atlas, whose graph is drawn schematically: a later paper B that cites an earlier paper A on the same task and benchmark and names a limitation of A. The segment becomes a frozen task. Its public package goes to the research agent, while the private package with the reference results, held-out labels and scorer bypasses the agent and reaches only the evaluator. The evaluator runs the agent's submission, recomputes the metric, compares the result with the starting point and the reference, and checks the claim behind it. The dashed path marks the loop that later versions close, in which a verified step becomes the next starting point.}
  \label{fig:pipeline}
\end{figure*}

The citation record already does the first of these jobs and much of the second. When a later paper on the same task and benchmark cites an earlier one and names a limitation of it, the literature has taken a step: the earlier method is a starting point, the limitation is a problem statement, and the later method is a hidden reference that solved the problem under that paper's own protocol. A chain of such papers yields a sequence of candidate tasks, each with an answer that the field already knows and the task can withhold. A builder turns a segment of the chain into a task and freezes its protocol, the agent receives the starting point and the limitation, and an evaluator that keeps the reference runs the submission (\cref{fig:pipeline}).

This report describes RSI-Atlas, the project that builds and tests this pipeline, and makes three contributions.
\begin{itemize}
  \item \textbf{The Paper Atlas} (\cref{sec:atlas}), a map of 7,317 papers from 28 venues. For 6,976 of them it holds structured records of what they evaluate, in which every entry cites a page and quotes the passage behind it. It groups 4,959 papers into 630 task--dataset clusters and traces 1,236 citation chains inside them, which show where candidate research steps lie.
  \item \textbf{A procedure that turns a chain segment into a task} (\cref{sec:tasks}): a limitation recorded with its source passage, a starting point that runs, a protocol frozen in advance, separate public and private packages, and an evaluator that runs every submission. The same material supports research, reproduction and claim-verification tasks.
  \item \textbf{A pilot on six chains} (\cref{sec:pilot}): 20 tasks in density estimation, graph learning, temporal action segmentation, language-model compression, semi-supervised learning and vision backbones, attempted by Codex running \gpt{} and by Claude Code running \opus. The six chains were built before the Atlas snapshot, so the pilot tests the task procedure rather than the Atlas.
\end{itemize}

The pilot answers the second question with a count and, more usefully, with a pattern. \gpt{} improved on the starting point in 12 tasks and \opus{} in 10, plus 2 more after the operators repaired the submitted code. Two features separate the steps the agents took from those they did not. On tasks that ask for a cheaper method at fixed or bounded quality, 16 of 18 submissions improved, two of them after repairs; these tasks started from implementations that had not been optimized for the targeted cost, and they held every candidate to the starting point's output or likelihood. On tasks that ask for a better method, 8 of the 18 comparable submissions improved, and the outcome tracked the size of the step that the task's reference shows: the agents improved in 7 of 8 submissions where the reference was more than 10\% better than the starting point and in none of 6 where a later paper was less than 4\% better (\cref{fig:steps}). The remaining improvement came on a task whose reference value was the starting value.

The pilot also answers part of the third question: a better metric does not by itself confirm a step. On the two chains where agents ran their own ablations, 3 of the 12 counted improvements came with an ablation that contradicted the proposed mechanism, and across the pilot 6 counted improvements took the same route as the task's reference. In a loop that draws its next limitation from the agent's analysis, a wrong explanation would steer the next round, so a verified step needs a second record beside the metric confirmed by execution: whether the agent's own controls support its explanation, and whether its route is new. A third finding is that a published step belongs to its protocol: under the task's budget, the starting models of two density tasks already did better than the last paper of their chain, so starting points and references have to be measured again in the protocol the loop runs in. The next version tests these patterns prospectively, on tasks fixed in advance and run several times per agent.

\Cref{sec:related} relates this design to existing work. \Cref{sec:atlas,sec:tasks} answer the first question, where tasks for research steps come from and how they are built. \Cref{sec:pilot} answers the second question, and \cref{sec:confirm,sec:lessons} answer the third: what execution confirms, what a claim record adds, and under which protocol the size of a step is defined. \Cref{sec:next} describes how later versions will repeat the step on its own output and extend the corpus to other sciences.

\section{Related work}\label{sec:related}

Five lines of work are closest to RSI-Atlas. They differ from it in what the agent must do, where the tasks come from and how outcomes are confirmed.

\paragraph{Reproducing published results.} PaperBench asks agents to replicate 20 ICML 2024 papers from scratch; each submission's reproduction script is rerun and graded by a language-model judge against 8,316 rubric items written with the papers' authors \citep{starace2025paperbench}. CORE-Bench builds 270 tasks from 90 papers; in each task an agent runs the authors' code and answers questions about its outputs \citep{siegel2024corebench}. ScienceAgentBench executes agent-written programs for 102 tasks drawn from 44 papers in four disciplines \citep{chen2025scienceagentbench}, and EXP-Bench builds 461 tasks from 51 papers, each giving the agent a research question, a method outline and a repository with key files masked \citep{kon2026expbench}. These benchmarks measure whether an agent can recover a known result. The reproduction mode of \cref{sec:taskunit} shares that goal, whereas the research mode hides the target and accepts any valid improvement.

\paragraph{Improving a starting solution.} MLE-bench packages 75 Kaggle competitions with local grading and their human leaderboards \citep{chan2025mlebench}. MLAgentBench provides 13 tasks with starter code and counts an improvement of at least 10\% over the baseline as success \citep{huang2024mlagentbench}, and MLGym provides 13 open-ended tasks with baselines \citep{nathani2025mlgym}. RE-Bench offers seven research-engineering environments, each with a starting solution and a reference solution that the agent never sees and that only normalizes the score \citep{wijk2025rebench}; it is the closest precedent for a hidden reference. On seven competition problems, MLRC-Bench scores how much of the gap between a baseline and the best human entry an agent closes \citep{zhang2025mlrcbench}, a normalization similar to the gap scores used on two of our chains (\cref{app:density,app:cora}). In these benchmarks a task is a competition or an environment with a given starting point, whereas our tasks start from a method on a citation chain and take their limitation from the later paper when that paper states one.

\paragraph{Mapping the literature.} The records of the Atlas belong to a line of work that extracts tasks, datasets, metrics and results from papers. \citet{hou2019identification} build leaderboards from such tuples, AxCell extracts results from the tables of machine-learning papers \citep{kardas2020axcell}, SciREX annotates tasks, datasets, metrics and methods and their relations across whole documents \citep{jain2020scirex}, and the Open Research Knowledge Graph stores research contributions as structured, comparable statements \citep{jaradeh2019orkg}. The Atlas keeps less of each paper, recording where each result is reported but not yet its value, and ties every entry to a page and a quoted passage; it uses the records for one purpose: to group papers that measure the same thing so that tasks can be built from them. A citation alone does not say why one paper cites another, a question studied as citation function or intent \citep{jurgens2018measuring,cohan2019structural}. The Atlas uses a citation only as evidence that the later paper knew the earlier one, and the task procedure establishes the limitation by reading the later paper (\cref{sec:taskunit}). Chain of Ideas also arranges papers into chains, as context from which an agent proposes new research ideas \citep{li2025chain}; in RSI-Atlas a chain supplies tasks with a hidden reference and an executed evaluation.

\paragraph{Tasks built from paper lineages.} Recent benchmarks have begun to hide a target paper. ResearchGym withholds the methods of five 2025 papers while keeping their datasets, evaluation scripts and baselines \citep{garikaparthi2026researchgym}. The Automated LLM Speedrunning Benchmark takes a chain of successive improvements from a competition log, the NanoGPT speedrun, and asks an agent to recover the next record from the script of the previous one \citep{zhao2025automated}. AI-Researcher gives agents the references of a target paper and compares their output with the hidden paper through language-model evaluators \citep{tang2025airesearcher}. Jr.\ AI Scientist starts from a baseline paper and its code, proposes improvements and writes a paper for reviewers \citep{miyai2026jr}, and ResearchClawBench ties 40 tasks to hidden published papers and scores each task with expert rubrics applied by a language-model judge \citep{xu2026researchclawbench}. LINEAGE builds graphs in which an earlier paper serves as an empirical baseline of a later one, but uses them for literature retrieval rather than task construction \citep{wu2026lineage}. To our knowledge, no prior benchmark combines the four elements of the task design in \cref{sec:tasks}: chains mined from the citation record, a limitation stated by the later paper, that paper as a hidden reference, and execution of every submission by an evaluator under a frozen protocol. \Cref{sec:pilot-chains} shows how fully each pilot task implements them.

\paragraph{Research systems and self-improvement.} The AI Scientist generates ideas, code, experiments and papers that an automated reviewer scores \citep{lu2024aiscientist,yamada2025aiscientistv2,lu2026towards}, and AlphaEvolve improves programs by keeping the changes that automated evaluators score well \citep{novikov2025alphaevolve}. The idea of recursive self-improvement goes back to Good's intelligence explosion \citep{good1965speculations} and to Gödel machines, which rewrite their own code once they can prove that the rewrite helps \citep{schmidhuber2007godel}. STOP applies a code-improving scaffold to itself while the language model stays fixed \citep{zelikman2024stop}, and the Darwin Gödel Machine replaces proofs with benchmark checks of each self-modification \citep{zhang2026darwin}. Whereas STOP and the Darwin Gödel Machine improve their own code and AlphaEvolve improves one program at a time, RSI-Atlas studies improvements to published research methods and builds the verified step that a loop over such methods would repeat.

\section{The Paper Atlas: where research steps come from}\label{sec:atlas}

\begin{figure}[t]
  \centering
  \includegraphics[width=\textwidth]{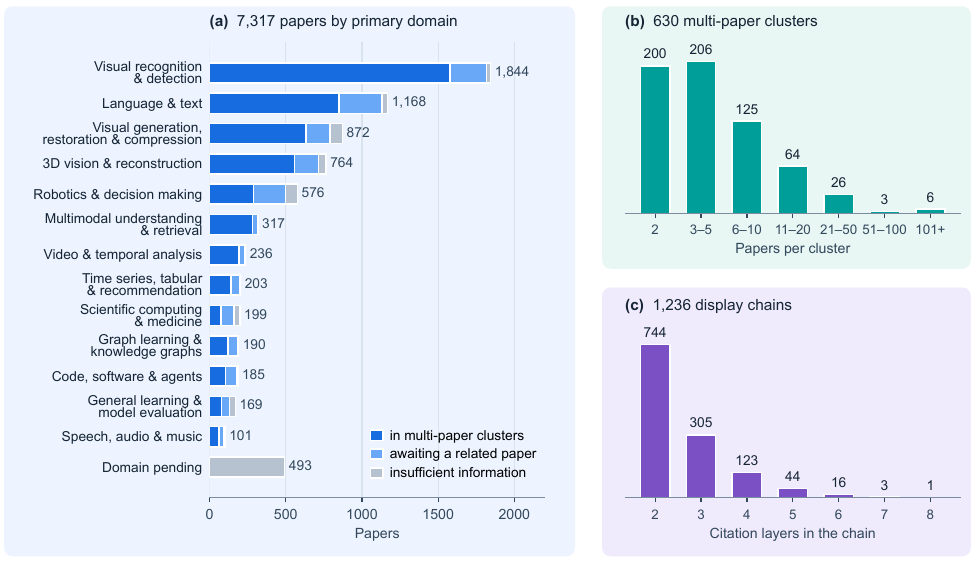}
  \caption{The Paper Atlas. (a) Papers per primary domain, split by clustering outcome; ``domain pending'' marks papers not yet assigned a primary domain. (b) Sizes of the 630 multi-paper clusters. (c) Lengths of the 1,236 display chains, the within-cluster chains that share no paper (\cref{sec:atlas-chains}), counted in citation layers; papers that cite each other share one layer.}
  \label{fig:atlas}
\end{figure}

A research task drawn from the literature needs a method to start from, a later method that improves on it, and an evaluation under which the two can be compared. The published record contains many such pairs, but they are spread across venues and years and reported in different ways. The Paper Atlas is our map of where they are. It records what each paper evaluates, groups papers that address the same task on shared evaluation data, and traces the citation paths inside each group. \Cref{fig:atlas} summarizes the snapshot used in this report, taken on 6 November 2026; the Atlas can be browsed at \url{https://rsi-atlas.ai}.

\subsection{Corpus and structured records}

The corpus contains 7,317 papers from 28 venues, led by CVPR (1,743 papers), NeurIPS (1,366), ICLR (1,060), ICCV (611), ICML (555) and ECCV (489); 91.7\% were published between 2019 and 2025. Every paper has a public code repository, a precondition for a starting point that runs (\cref{sec:startpoint}). Every PDF has a recorded source (proceedings sites for 4,677 papers, arXiv for 2,604, and author pages, code repositories or institutional archives for 36) and a recorded SHA-256 hash, so that evidence can point to a page of an exact file. For 6,861 of the 6,976 extracted papers, the file that was read matches the recorded hash; for the other 115 it is a later arXiv version of the same paper.

From the PDFs we extract structured records in which every entry carries a page number and a quoted passage that supports it. A record lists the tasks the paper evaluates, with benchmark, metrics and, where the paper states it, the split; the models it trains; the baselines it compares against; and its references. Two language models performed the extraction on text read with PyMuPDF: GPT-6-Sol, a different model from the \gpt{} agent of the pilot, for 4,491 papers, and \opus{} for 2,485. The 6,976 records hold 147,787 evaluation entries (supported by 356,236 quoted passages), 148,765 baseline entries and 397,068 references; the median paper has 13 evaluation entries and 53 references. The extraction is a first pass. Of the 6,976 records, 6,220 are marked partial, most because their completeness across text chunks has not yet been reviewed and most of the others because the paper refers to appendices or supplementary material that the extraction did not see. So far 9.7\% of the baseline entries are linked to a source paper, too few to require at scale that a later paper report the earlier one as a baseline, the optional rule of \cref{sec:taskchains}. Any fact that a task depends on is therefore checked again against the source pages when the task is built (\cref{sec:taskunit}).

\subsection{Primary domains and task--dataset clusters}

Each paper is assigned one primary domain and one primary task by rules applied to its evaluation entries, so that every paper is counted once (\cref{fig:atlas}a). Thirteen domains cover 6,824 papers; the other 493 are still pending a domain. Vision dominates the corpus: recognition, generation, 3D vision and video together account for 3,716 papers, 50.8\% of the total.

A domain is too broad to define a research step, so papers are grouped by what they measure. Two papers are linked when they have the same primary task, the same explicit subtask and setting (for example semi-supervised or federated learning), and at least one family of evaluation datasets in common. The strength of a link is the number of shared datasets divided by the geometric mean of the two papers' dataset counts. Louvain community detection \citep{blondel2008fast}, with resolution 1 and seed 7317, then partitions each connected component. A single bridge edge between two groups of at least two papers each is cut, so that one link cannot join two otherwise separate communities. No title words, embeddings or target cluster sizes enter the procedure.

The result is 630 multi-paper clusters containing 4,959 papers, 67.8\% of the corpus (\cref{fig:atlas}b). The median cluster has four papers and the largest has 212; in 397 clusters at least one dataset is shared by every member. A further 1,482 papers have usable records but are still awaiting a related paper under these rules. The remaining 876 lack the information needed: 341 have no extraction, 383 have no evaluation dataset that can be matched across papers, 136 evaluate tasks that the current rules do not yet cover, 15 report no original evaluation, and one has an empty evaluation list. Re-running with seeds 17 and 2026 changes the partition little (adjusted Rand index 0.967 and 0.945, respectively, relative to the reported partition), whereas the resolution sets the granularity: resolutions of 0.75 and 1.25 give 531 and 798 clusters, respectively.

\subsection{Citation chains inside clusters}\label{sec:atlas-chains}

A cluster shows that its papers measure the same thing, and a citation shows that the later paper knew the earlier one. We write $A \rightarrow B$ when $B$ cites $A$, using the extracted reference lists matched to papers in the corpus. A layer is one paper, or one group of papers that cite each other, often concurrent work. The 630 clusters contain 6,917 such citation edges, and 3,830 of the 4,959 clustered papers take part in at least one.

Two views summarize these edges. The first counts branching paths: every maximal directed path from a paper that cites no earlier paper in its cluster to one that no later paper in the cluster cites. A paper can lie on many such paths. There are 376,103 within-cluster paths, the longest spanning 19 citation layers, and on 369,016 of them a dataset is shared by every paper on the path. The second view, the display chains, covers the citation graph with the fewest paths that share no paper and keeps those with at least two layers, so each paper appears in at most one display chain. These 1,236 display chains, the citation chains that the Atlas shows, hold 3,279 papers in two to eight layers (\cref{fig:atlas}c), and 1,142 of them share at least one dataset across all members. Branching paths measure how much citation structure the clusters hold; display chains are what a task builder browses.

\section{From a chain segment to a task}\label{sec:tasks}

A display chain from the Atlas is a candidate, not a task. A citation may acknowledge background rather than respond to a limitation, and a shared dataset does not guarantee a shared protocol. A chain segment, a pair or short sequence of consecutive papers on a chain, becomes a task whose outcome can be trusted once six components are in place: stricter chain rules (\cref{sec:taskchains}), a limitation that an experiment can show (\cref{sec:taskunit}), a starting point that runs (\cref{sec:startpoint}), a protocol frozen in advance (\cref{sec:protocol}), separate public and private packages (\cref{sec:packages}), and an evaluator that runs the submission (\cref{sec:evaluation}). Three roles recur. The builder constructs a task and freezes its protocol. The evaluator runs each submission, and a separate scorer computes the metric from its outputs. The operators run the infrastructure, and any repair they make to submitted code is labeled.

\subsection{Task chains}\label{sec:taskchains}

Chains used for tasks follow stricter rules than display chains. Each link $A \rightarrow B$ must be a direct citation found in the bibliography of $B$, recorded with its page. Every paper on the chain must evaluate the same task on a confirmed shared benchmark, that is, the same dataset and not merely a dataset with the same name. A chain is kept only if it cannot be extended at either end under that benchmark; branches become separate chains. An optional stricter rule further requires that $B$ report $A$ as an experimental baseline on the shared benchmark. These rules establish that $B$ knew $A$ and measured the same thing. They do not show that $B$ removed a limitation of $A$; \cref{sec:taskunit} records the limitation and \cref{sec:protocol} tests it.

\subsection{What a task contains}\label{sec:taskunit}

A task is built around a target paper $B$ on a chain and the prior work that $B$ discusses. The builder reads the introduction and related-work sections of $B$, finds a limitation that $B$ attributes to the prior work, and records it with the quoted passage, its page and the hash of the source file. A limitation enters a task only if an experiment can show it; a general remark that a method leaves room for improvement does not qualify. Each task holds one limitation, the prior methods it concerns and one experimental protocol.

The same material supports three modes, which differ in what the agent sees and in what counts as success (\cref{tab:modes}). In the research mode, the default, the agent receives the prior work, the limitation, the data and the constraints, and must find an improvement. The target paper stays hidden as a reference showing that the problem can be solved. A different method that meets the constraints is an equally valid answer, so the score ignores both the name of the submitted method and its resemblance to the target paper. Hiding the target paper keeps it out of the task package, not out of the models' training data, so the research mode asks whether an agent reaches an improvement under a frozen protocol and budget by any route, including recalling a method it already knows. In the reproduction mode the agent receives the target method and must reproduce its result. In the claim-verification mode the agent must establish under which conditions the stated limitation holds, and a well-supported finding that it does not hold under the protocol is a valid outcome.

\begin{table}[t]
  \centering
  \small
  \caption{Three task modes built from the same chain segment. The target paper is the later paper of the segment.}
  \label{tab:modes}
  \begin{tabularx}{\textwidth}{@{}l>{\raggedright\arraybackslash}X>{\raggedright\arraybackslash}X>{\raggedright\arraybackslash}X@{}}
    \toprule
    Mode & The agent receives & Success means & The target paper is \\
    \midrule
    Research (default) & the prior work, the limitation, the data, the constraints & a reproducible improvement supported by experiments & a hidden reference, not the only correct answer \\
    Reproduction & the target method, its setup, the permitted code & the target result, reproduced & given to the agent as the method to reproduce \\
    Claim verification & the claim, related methods, the protocol & the conditions under which the limitation holds & the source of the claim \\
    \bottomrule
  \end{tabularx}
\end{table}

\subsection{A starting point that runs}\label{sec:startpoint}

An agent can only improve a method that runs. For each chain, the builder pins every author repository to a commit, keeps compatibility patches apart from the original code, builds one container image with locked dependencies, and provides an entry point for each paper and a common one for the chain. Before any agent sees a task, its starting point should complete a short run on real data; the run covers training, evaluation, saving a checkpoint and resuming from it. The environment package of the eleven-paper vision-backbone chain in the pilot shows what this takes: its image pins 117 dependencies, all eleven models passed GPU forward and backward checks inside the container, and each model completed a short training run on a 512-image ImageNet subset, followed by evaluation, checkpointing and resumption in a new process.

\subsection{A protocol frozen in advance}\label{sec:protocol}

The builder then freezes everything the comparison depends on: the data version and its hash, the splits, the model and its starting weights, the training budget and seeds, the metric and its direction, the resource limits, and the rules for failed runs and exhausted budgets. The starting and reference values of a task should come from runs under this protocol, whether in the chain's reproduction package, in runs made for the task, or in controls timed in the same session as a submission. Before release these runs serve three checks: whether the limitation appears under the protocol, whether the reference fits the budget, and how much results vary, which sets the tolerance used in scoring. A task whose limitation does not appear is revised, converted to claim verification, or dropped.

\subsection{Public and private packages}\label{sec:packages}

Each task is split into a public package and a private one. The public package holds what the agent needs to work: the background and the limitation, the starting code, the data interface, the metric, the budget and the submission format. The private package holds the hidden reference and its results, held-out labels, the scorer and the review records. The agent works in an isolated environment that contains only the public package, so it can read neither the private package nor the builder's notes nor another agent's work. When a task has a hidden reference, the public text is checked before release for leaks of the target paper's title, method name, authors, repository and solution hints. Every model call, tool call, code revision, failed attempt and final submission is recorded as an observable trajectory.

\subsection{Independent evaluation}\label{sec:evaluation}

The evaluator does not use the numbers that the agent reports. It runs the sealed submission under the frozen protocol, where the submitted code cannot read the private package, and a separate scorer computes the metric from the outputs and the held-out labels. Validity is decided first: a submission that cannot be executed, uses forbidden information, changes the protocol or exceeds its budget is marked invalid before its quality is assessed. For each task the evaluator then reports the native metric, such as F1, perplexity, accuracy or runtime, beside the values of the starting point and the reference, so that a verified research step reads directly as an improvement over the starting point.

Results that needed help carry a label. An unassisted result was obtained within the attempt's original budget and without repairs. A recovered result was completed in an additional engineering window after the original budget ran out. An assisted result followed a repair of the agent's code by the operators; the agent's own attempt is kept beside it as its unassisted result. Comparisons at equal budget use only unassisted results.

\section{A pilot on six chains}\label{sec:pilot}

The pilot tests whether the procedure of \cref{sec:tasks} yields tasks that current agents can work on and whose outcomes an evaluator can confirm. This section reports in how many tasks the evaluator confirmed an improvement over the starting point, which kinds of step the agents took, and what running the submissions revealed. \Cref{app:chains} documents every task.

\subsection{Chains and tasks}\label{sec:pilot-chains}

\Cref{tab:chains} lists the six chains, which come from six areas of machine learning and contain six to eleven papers each. They predate the Atlas snapshot: each was built from direct citations within a candidate set of papers for its area, and each comes with a reproduction package that pins the papers and their author code. The pilot therefore tests the task procedure rather than the Atlas. In five chains every paper evaluates on the shared benchmark; the semi-supervised chain has no dataset shared by all seven papers, and its tasks use CIFAR-10.

\begin{table}[tbp]
  \centering
  \small
  \caption{The six chains of the pilot. $A \rightarrow B$ means that $B$ cites $A$; the number of papers on each chain is given in parentheses.}
  \label{tab:chains}
  \begin{tabularx}{\textwidth}{@{}l>{\raggedright\arraybackslash}Xlc@{}}
    \toprule
    Area & Chain & Data & Tasks \\
    \midrule
    Density estimation & MAF $\rightarrow$ TAN $\rightarrow$ NAF $\rightarrow$ B-NAF $\rightarrow$ NSF $\rightarrow$ LRS (6) & MiniBooNE & 5 \\
    Graph learning & GCN $\rightarrow$ GAT $\rightarrow$ DropEdge $\rightarrow$ GRAND $\rightarrow$ NAGphormer $\rightarrow$ Classic GNNs (6) & Cora & 3 \\
    Action segmentation & MS-TCN $\rightarrow$ ASRF $\rightarrow$ ASFormer $\rightarrow$ UVAST $\rightarrow$ DiffAct $\rightarrow$ ActFusion (6) & 50 Salads & 3 \\
    LLM compression & SmoothQuant $\rightarrow$ SparseGPT $\rightarrow$ LLM-Pruner $\rightarrow$ Wanda $\rightarrow$ DSnoT $\rightarrow$ OmniQuant $\rightarrow$ PB-LLM $\rightarrow$ BiLLM $\rightarrow$ SVD-LLM (9)$^{\S}$ & WikiText-2 & 3 \\
    Semi-supervised learning & Mean Teacher $\rightarrow$ Label Propagation $\rightarrow$ Noisy Student $\rightarrow$ FixMatch $\rightarrow$ DARP $\rightarrow$ CReST $\rightarrow$ CoSSL (7) & CIFAR-10$^{\dagger}$ & 3 \\
    Vision backbones & ViT $\rightarrow$ T2T-ViT $\rightarrow$ PVT $\rightarrow$ Swin $\rightarrow$ Vision Longformer $\rightarrow$ NesT $\rightarrow$ RegionViT $\rightarrow$ CrossFormer $\rightarrow$ EfficientFormer $\rightarrow$ SwiftFormer $\rightarrow$ RepViT (11) & ImageNet-1K$^{\ddagger}$ & 3 \\
    \bottomrule
  \end{tabularx}\\[3pt]
  \parbox{\textwidth}{\footnotesize\raggedright $^{\S}$ A branch DSnoT $\rightarrow$ SLEB adds a tenth paper. $^{\dagger}$ No dataset is shared by all seven papers; the tasks use CIFAR-10. $^{\ddagger}$ The benchmark of the chain; its operator tasks use fixed random inputs.}
\end{table}

Twenty tasks were built from these chains, 17 in the research mode and 3 in the claim-verification mode. The claim-verification tasks ask the agent to test the stated limitation and then reduce the cost it names, and \cref{tab:results} reports that cost. Nine cost tasks target runtime, memory or parameter count at fixed or bounded quality, and eleven quality tasks target likelihood, accuracy, F1, perplexity or cross-entropy.

The tasks implement the design of \cref{sec:tasks} to different degrees (\cref{tab:design} in \cref{app:chains}). Ten take a later paper on the chain as the hidden reference. In four the reference is the starting method as run in the reproduction package: its own result for OmniQuant, SVD-LLM and graph construction, and for label propagation the full two-stage method, whose first stage is the starting point. In three the reference is a model or implementation prepared for the task. Below, `reference' covers all three kinds, each kept in the private package, and `hidden reference' means a later paper. The three operator tasks have none; the authors' optimized code was part of their public packages, and in the Longformer and RepViT tasks it is timed as a control. Where the later paper states a limitation, the task quotes it; for several tasks, among them the operator tasks and two action-segmentation tasks, the builders formulated the limitation from the method or the starting code. Many starting and reference values came from the reproduction packages, and \cref{sec:lessons} shows how far they hold under the task protocols.

\subsection{Agents and execution}\label{sec:pilot-setup}

The agents were Codex \citep{openai2026codex} running \gpt{} and Claude Code \citep{anthropic2026claudecode} running \opus{}. We write \gpt{} and \opus{} for the two systems (model identifiers \nolinkurl{openai/gpt-6-astra} and \nolinkurl{anthropic/claude-opus-5}); each result reflects the model and its client together. The agents never saw a private package, the builders' notes or each other's work.

Two execution regimes were used. On the MiniBooNE and Cora chains the agents worked in stages and could launch a limited number of jobs through the task's research harness, for example at most 28 jobs and 12,000 optimizer steps for a MiniBooNE task, or two complete training jobs for the Cora depth task; the evaluator then retrained or re-ran each sealed submission. On the other four chains the agents wrote code without running it, which kept every execution inside the evaluator's runner and budget. In each call the agent received the shared rules of the pipeline, a participant instruction and a catalog of the public files; it requested files by path and returned complete source files, with no shell, search or other tools. For each task it could make up to 64 model calls within 14,400~s (4~h). This code-writing budget is separate from the evaluation window in which the runner later trained and tested the submission, and for action segmentation the code was written and sealed separately for each fold. No test or reference value was returned to the agents in this regime; when a run failed, the agent could receive the error trace and revise its code within the same budget. In both regimes a separate scorer computed every metric from the outputs. All work ran on RTX 4090 GPUs in isolated workers confined by Landlock, with no network access other than to the model gateway; on the Cora chain and the four code-only chains, each worker also ran under seccomp filters and a separate user account.

\subsection{How many steps the agents took}\label{sec:results}

\Cref{tab:results} gives, for every task, the starting value, the reference, the gap between them and the value of each agent's submission. To summarize across tasks whose metrics differ, we count a submission as an improvement when it meets the task's validity constraints, such as the likelihood margin of the density tasks, and beats the starting value by more than the threshold set in the task package or, for a task without one, by more than 1\% of the starting value. Replacing the 1\% default by any value between 0.7\% and 5.9\% gives the same counts. At 10\%, the threshold of MLAgentBench \citep{huang2024mlagentbench}, two results drop out: \gpt{} on label propagation, an improvement of 9.99\%, and \opus{} on SVD-LLM, a 5.9\% reduction in perplexity.

\begin{table*}[tbp]
  \centering
  \footnotesize
  \caption{Results of the pilot. Submissions are measured by the evaluator under each task's frozen protocol; the starting value (Start) and the reference (Ref.) come from the chain's reproduction package or from runs made for the task (\cref{app:chains}). Gap is the improvement of the reference over the starting value, as a percentage of the starting value. Values are means over seeds for density estimation and graph learning, means over five folds for action segmentation, a geometric mean over three seeds for graph construction, and single evaluation runs elsewhere. Bold marks an improvement over the starting value under the counting rule of \cref{sec:results}. NLL: negative log-likelihood in nats; W3A16: 3-bit weights and 16-bit activations; F1@50: segment-level F1 at 50\% overlap; $G$: speed-up over the starting operator.}
  \label{tab:results}
  \setlength{\tabcolsep}{4pt}
  \begin{tabular}{@{}llrrrrr@{}}
    \toprule
    Task & Metric & Start & Ref. & Gap (\%) & \gpt & \opus \\
    \midrule
    \multicolumn{7}{@{}l}{\textit{Density estimation on MiniBooNE}} \\
    Efficiency$^{c}$ & inverse latency, batch 64 (ms) $\downarrow$ & 21.1 & --- & --- & \textbf{16.27} & \textbf{3.56} \\
    Capacity & held-out NLL $\downarrow$ & 20.771 & 18.526 & 10.8 & \textbf{16.111} & \textbf{16.681} \\
    Compactness$^{c}$ & parameters (thousands) $\downarrow$ & 547.9 & 30.4$^{\dagger}$ & --- & \textbf{279.6} & 106.6$^{m}$ \\
    Inversion$^{c}$ & inverse latency, batch 64 (s) $\downarrow$ & 2.711 & --- & --- & \textbf{1.031} & \textbf{0.217} \\
    Coupling & held-out NLL $\downarrow$ & 25.297 & 17.442 & 31.1 & 25.163 & \textbf{18.499} \\
    \midrule
    \multicolumn{7}{@{}l}{\textit{Graph learning on Cora}} \\
    Depth & cross-entropy (nats) $\downarrow$ & 1.620 & 1.562 & 3.6 & 1.640 & 1.661 \\
    Memory & peak memory (MiB) $\downarrow$ & 484.7 & 53.2 & 89.0 & \textbf{422.0} & \textbf{56.1}$^{a}$ \\
    Inference & latency (\textmu s) $\downarrow$ & 656.3 & 382.0 & 41.8 & \textbf{391.6} & \textbf{385.8}$^{a}$ \\
    \midrule
    \multicolumn{7}{@{}l}{\textit{Temporal action segmentation on 50 Salads}} \\
    A: boundaries & F1@50 (\%) $\uparrow$ & 64.58$^{*}$ & 74.67 & --- & 60.04 & 60.22 \\
    B: long-range context & F1@50 (\%) $\uparrow$ & 74.67$^{*}$ & 79.48 & --- & 73.88 & 72.32$^{r}$ \\
    C: conditioning & F1@50 (\%) $\uparrow$ & 83.81 & 84.81 & 1.2 & 84.37 & 82.40$^{r}$ \\
    \midrule
    \multicolumn{7}{@{}l}{\textit{Language-model compression on WikiText-2}} \\
    Wanda, 60\% sparsity & perplexity $\downarrow$ & 10.705 & 10.678 & 0.3 & 11.737 & 12.416 \\
    OmniQuant, W3A16 & perplexity $\downarrow$ & 6.503 & 6.503 & 0 & 6.514 & 6.595$^{r}$ \\
    SVD-LLM whitening & perplexity $\downarrow$ & 7.892 & 7.892 & 0 & 7.890 & \textbf{7.424} \\
    \midrule
    \multicolumn{7}{@{}l}{\textit{Semi-supervised learning on CIFAR-10}} \\
    Mean Teacher, 40 labels & accuracy (\%) $\uparrow$ & 23.88 & 92.99 & 289.4 & \textbf{34.23} & \textbf{76.34} \\
    Label propagation & accuracy (\%) $\uparrow$ & 79.37 / 79.02 & 88.38 & 11.4 / 11.8 & \textbf{87.30} & \textbf{88.61} \\
    Graph construction & speed-up $G$ $\uparrow$ & 1 & 1 & 0 & \textbf{8.26} & \textbf{95.48} \\
    \midrule
    \multicolumn{7}{@{}l}{\textit{Vision backbones: operator tasks}} \\
    Swin windows & speed-up $G$ $\uparrow$ & 1 & --- & --- & \textbf{3.09} & 1.00 \\
    Longformer attention & speed-up $G$ $\uparrow$ & 1 & --- & --- & \textbf{1.69} & \textbf{2.77} \\
    RepViT fold & speed-up $G$ $\uparrow$ & 1 & --- & --- & \textbf{6.65} & \textbf{11.27} \\
    \bottomrule
  \end{tabular}\\[3pt]
  \parbox{\textwidth}{\footnotesize\raggedright $^{c}$ claim-verification task; all other tasks are research tasks. ${}^{a}$ assisted result, after a repair by the operators. ${}^{r}$ recovered result, completed in an additional engineering window. ${}^{*}$ starting value measured under a different protocol. ${}^{m}$ fails the task's likelihood margin. ${}^{\dagger}$ the reference reaches this size only with a held-out NLL of 24.21, outside the task's likelihood margin, so no gap is given. ---: no value on this metric. For label propagation, Start is the accuracy at the end of the first stage in each agent's run.}
\end{table*}

\gpt{} improved on every cost task and on three quality tasks; \opus{} improved on five cost tasks unassisted and two more after repairs, and on five quality tasks (\cref{tab:counts}). The four submissions on action-segmentation tasks A and B are left out because their starting values were measured under a different protocol, and the remaining 12 of the 40 submissions did not improve on the starting point. On individual tasks the two agents differ widely, in both directions: with 40 labels \opus{} reached 76.34\% where \gpt{} reached 34.23\%, while on the Swin operator the \gpt{} kernel ran 3.09 times as fast as the starting path and the \opus{} code fell back to it. Each task--agent pair produced one scored result, so these differences describe two submissions rather than two agents.

\begin{table}[tbp]
  \centering
  \small
  \caption{Submissions that improved on the starting value, by agent and task type, under the counting rule of \cref{sec:results}. Quality tasks exclude action-segmentation tasks A and B, whose starting values were measured under a different protocol; All tasks counts their four submissions as not improved. The row for both agents includes the two assisted results of \opus.}
  \label{tab:counts}
  \begin{tabular}{@{}lccc@{}}
    \toprule
     & Cost tasks & Quality tasks & All tasks \\
    \midrule
    \gpt & 9 of 9 & 3 of 9 & 12 of 20 \\
    \opus & 5 of 9, plus 2 assisted & 5 of 9 & 10 of 20, plus 2 assisted \\
    \midrule
    Both agents & 16 of 18 & 8 of 18 & 24 of 40 \\
    \bottomrule
  \end{tabular}
\end{table}

A reference marks a step known to be possible under the protocol, not a ceiling. On MiniBooNE capacity both agents reached a lower held-out NLL than the hidden reference, NAF, and than LRS, the last paper of the chain, with NAF and LRS trained for the same 1,000 steps as the submissions. On label propagation \opus{} reached 88.61\%, compared with 88.38\% for the original method under the same selection rule, the best test accuracy during training. On the Cora inference task both submissions came within 3\% of the reference latency on average over the two seeds, and on the RepViT operator the \opus{} kernel ran 1.68 times as fast as the authors' fusion timed in the same run, with the \gpt{} submission on par with that fusion.

\subsection{Which steps the agents took}\label{sec:which}

\begin{figure*}[tbp]
  \centering
  \includegraphics[width=\textwidth]{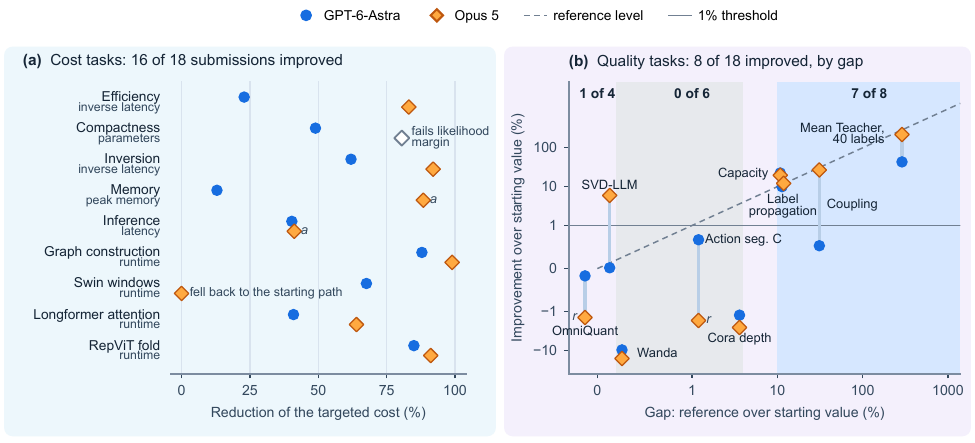}
  \caption{Which steps the agents took, with the values of \cref{tab:results}. (a) Cost tasks: reduction of the targeted cost relative to the starting point, computed as one minus the ratio of submission to starting value for latency, memory and parameters, and as the time-saving score of \cref{eq:timesaving} for speed-ups. (b) Quality tasks: improvement over the starting value, in the direction of each metric, against the gap by which the task's reference improves on the starting value. Both axes of (b) are linear within $\pm 1\%$ and logarithmic beyond; the dashed line marks an improvement equal to the gap, and the solid line the 1\% counting threshold. The shaded groups follow \cref{sec:which}: a later paper less than 4\% better (gray) and a reference more than 10\% better (blue); OmniQuant and SVD-LLM, whose reference equals the starting value, sit at a gap of zero and are drawn slightly apart. Bold counts give the submissions that improved in each group. Labels $a$ and $r$ mark the assisted and recovered results of \opus{}. Action-segmentation tasks A and B are omitted because their starting values were measured under a different protocol.}
  \label{fig:steps}
\end{figure*}

The improvements are not spread evenly over the tasks. On the nine cost tasks, 16 of the 18 submissions improved on the starting point, two of them after repairs; both exceptions come from \opus{}: a compact density model that failed the likelihood margin and a Swin submission whose code fell back to the starting path. On the nine quality tasks that can be compared with the starting point, 8 of the 18 submissions improved. The contrast holds in both regimes: on the two chains where agents could run jobs, 9 of 10 cost submissions and 3 of 6 quality submissions improved, and on the four code-only chains, 7 of 8 and 5 of 12 (\cref{fig:steps}).

Two properties of the cost tasks help explain the contrast. Every cost task holds a candidate to the starting point's output or likelihood, exactly or to a tight tolerance for the operator tasks, the graph construction and the two Cora tasks, and within a likelihood margin for the density tasks, so whether a candidate is correct is settled by a check rather than by comparing noisy training runs. And the starting implementations had not been optimized for the targeted cost, which left room for familiar engineering moves: fusing the operators of a block, moving an exact nearest-neighbor search to the GPU, replaying a captured CUDA graph, or replacing a framework optimizer with an explicit one. Graph construction shows how much room there was. Its reference is the starting code itself, yet the two submissions built the same exact graph 8.26 and 95.48 times as fast. The contrast does not rest on the operator tasks, whose public packages contained the authors' optimized code: on the other six cost tasks, 11 of 12 submissions improved, two of them after repairs. A quality step has neither advantage: no check certifies it, and it has to beat the starting point on a measured quality that in most of these tasks depends on training runs whose outcome varies.

Among quality tasks, the outcome tracked the size of the step that the task's reference shows under the protocol, measured by the gap between the starting point and the reference (\cref{fig:steps}b). Where the reference was more than 10\% better than the starting point (MiniBooNE capacity and coupling, Mean Teacher with 40 labels, and label propagation), 7 of the 8 submissions improved. Where a later paper was less than 4\% better (Cora depth, action-segmentation task C and Wanda), none of the 6 did. The split holds in both regimes: 3 of 4 compared with 0 of 2 on the chains where agents could run jobs, and 4 of 4 compared with 0 of 4 on the code-only chains. A small gap means that the stated limitation barely limits the metric under the protocol, which is the condition the check of \cref{sec:protocol} looks for before release. Wanda shows the consequence. \opus{} cut the per-layer pruning error by 20.7\% across all 224 projections, yet perplexity rose from 10.705 to 12.416, the outcome named by the falsification test that \opus{} had stated before the run, while the reference itself lowers perplexity only from 10.705 to 10.678. On OmniQuant and SVD-LLM the reference is the package's result for the starting method, so the gap is zero by construction; \opus{} still lowered the SVD-LLM perplexity by 5.9\%.

Both patterns say where a loop of research steps can start: with steps whose correctness can be checked against the starting point and verified in a single evaluation, and, among quality steps, with those whose reference shows a large step. \Cref{sec:next} describes how the next version tests both patterns.

\subsection{What execution confirms, and what the claim record adds}\label{sec:confirm}

Running every submission catches what a report cannot. The Swin submission of \opus{} contained a fused kernel, but its self-check computed gradients where gradients are disabled, failed on every input and silently selected the starting path, so the kernel never ran; the agent, which could not run code, had no way to see this. Fallbacks can also come from the environment: five evaluation runs of the \opus{} RepViT submission took a slower path because of build faults before the submitted kernel ran. Where results were reproduced, they matched: all ten MiniBooNE results passed a fresh-training replay, fresh retraining matched the Cora depth results to within $5 \times 10^{-7}$, and rescoring the saved predictions of all 30 action-segmentation fold runs gave every score exactly.

A better metric, however, does not confirm the account that comes with it. The MiniBooNE rubric and the Cora evidence items score the agent's claim beside its metric. On these two chains, where agents ran their own ablations, 3 of the 12 counted improvements came with an ablation that went against the proposed mechanism. The \opus{} coupling model lowered the held-out NLL from 25.297 to 18.499, but its spline transform did worse than its own affine ablation; the \gpt{} efficiency submission cut the inverse time from 21.1 to 16.27~ms, but its own ablation, without the proposed change, was faster still; and the \gpt{} inversion submission cut the inverse time from 2.711 to 1.031~s, while the speed-up it had predicted for Newton's method over its own bisection ablation did not appear at batch 64. All three agents state the contradiction in their reports. The assisted \opus{} memory result went wrong the other way: its ablation supported the proposed mechanism, but the report read it backwards, and the operators corrected the reading (\cref{app:cora}). An ablation can also confirm a claim: removing the shared block structure from the \opus{} efficiency model lengthened its chain of sequential conditioner calls from 16 to 172 and raised the batch-64 inverse latency by a factor of 10.9.

Six counted improvements took the same route as the task's reference. Four followed a reference that the builders had prepared or the original method of the task: the \opus{} memory submission replaced PyTorch with a NumPy and SciPy engine, as the prepared reference does; both inference submissions replayed the forward pass as a captured CUDA graph, as the prepared reference does, the \opus{} one together with a fused attention layer; and the \opus{} label-propagation submission propagated labels over a nearest-neighbor graph of the current embeddings, the core of the original method. Two used the method of the hidden later paper: the \opus{} inversion model is an autoregressive flow with analytically invertible rational-quadratic splines, the method of NSF, and with 40 labels \gpt{} combined hard pseudo-labels on strongly augmented images above a confidence of 0.95 with seven times as many unlabeled images per step, the recipe of FixMatch. The task design accepts such convergence (\cref{sec:taskunit}). Four of the five submissions whose reference has a value on the task's metric came within 6\% of it or went past it. The \gpt{} 40-label recipe did not: it reached 34.23\%, compared with 92.99\% for FixMatch, so a recipe that resembles the reference does not by itself reproduce its result.

A verified step therefore needs two records. The first is the metric, confirmed by running the submission under the frozen protocol. The second is the claim record: whether the agent's own controls support its explanation, and whether its route is new or the reference's own. Both matter for a loop. A step whose explanation is wrong would hand the next round a wrong limitation, and a step that follows the reference's route shows that an agent can repeat known progress, not yet that it can extend it.

\section{What the pilot shows about verifying research steps}\label{sec:lessons}

The claim record of \cref{sec:confirm} concerns what a submission says about itself. The pilot also shows that the values a step is measured against need the same care as the step itself.

\paragraph{A published step holds under its own protocol.} A reference shows that a limitation can be overcome, but only under the conditions in which it was measured, and moving a step into a task's protocol changed several of the steps on these chains. Under the MiniBooNE budget of 1,000 training steps, the dense starting models of the compactness and inversion tasks reached a held-out NLL of 16.64, better than the 17.44 of LRS, the last paper of the chain, trained with the same budget; the compactness reference, a B-NAF variant, reaches its 30,445 parameters only at an NLL of 24.21. On the language-model chain, the step from Wanda to DSnoT shrinks under the task's protocol to a change in perplexity from 10.705 to 10.678 (\cref{sec:which}). Other starting values came from settings that differ from the task's: the package's values for action-segmentation tasks A and B were measured on all frames and under a different training protocol, respectively, and the starting Mean Teacher run drew one unlabeled batch per labeled batch, whereas the FixMatch reference drew seven. The size of a step therefore belongs to the protocol in which it is measured, not to the paper that reported it. For a loop, this means that starting points and references are measured again under the loop's own protocol before a task is released, which is what the checks of \cref{sec:protocol} do.

\paragraph{A speed-up holds under one configuration.} The operator tasks measure a submission against the starting operator timed in the same session, and the pilot shows why that matters. The starting Longformer operator ran in 8.08~ms in one agent's session and in 6.62~ms in the other's, and the RepViT measurements used four CPU threads for \gpt{} and a single thread for \opus{}. Differences of this size do not change whether these submissions are faster than the starting operator, but they do change any comparison between the two agents. The time-saving score (\cref{eq:timesaving}) also gives zero both to a correct submission that is no faster and to one that fails its checks. Timing tasks therefore measure every candidate under one fixed configuration and report the raw timings beside the score.

\paragraph{Claims need a judge outside the comparison.} Scoring the claim record needs a judge. On the Cora chain, 60 of the 100 points of each score come from 24 binary evidence items, such as a registered plan, an executed ablation or a saved per-seed ledger, and a separate instance of Codex running \gpt{} judged them, a model that also competed. The counts of \cref{sec:results} use the native metrics alone. A judge should not share a model with a contestant: evidence items are best checked by deterministic rules where possible and otherwise by a model that is not under test.

\section{Toward recursive self-improvement and other sciences}\label{sec:next}

This report evaluates the single step: a task built from a chain segment, an agent working on it, and an evaluator that decides whether the result improves on the starting point. Recursive self-improvement needs that step to repeat on its own output. The next versions move toward this in three ways: they draw tasks from the Atlas and fix them in advance, they close the loop, and they widen the corpus beyond machine learning.

\paragraph{Tasks from the Atlas, fixed in advance.} The next version will start from the display chains of the Atlas, apply the task-chain rules of \cref{sec:taskchains}, and settle the starting and reference values, the scoring rule and the resource budget before any agent runs, so that every score is prospective. Each task will be run several times per agent, so that two agents can be compared on the same task and the patterns of \cref{sec:which} can be tested prospectively. The size of the published step is the natural variable for that test. The Atlas records already point to the passage or table that reports each result; extracting the reported values would let a builder screen segments by their published gap before calibrating the remaining ones under the task's protocol (\cref{sec:protocol}).

\paragraph{Closing the loop.} When a step is verified, the improved method can become the starting point of the next task in the same line of work. The next limitation comes from the next paper on the chain or from the analysis in the agent's own report; the second source is where the claim record of \cref{sec:confirm} is used. Where the chain has a next paper, its method serves as the reference, but that paper improved on $B$ rather than on the agent's method, so its limitation and its result are checked against the new starting point under the protocol before the round begins. Two quantities then describe progress: how many consecutive verified steps an agent achieves before it stops improving, and the cost of each step. The evaluator, the data and the budget stay fixed across rounds and outside the agent's reach, because a step that is not checked independently carries its error into every step built on it.

\paragraph{Beyond machine learning.} The pipeline needs evaluations that can be compared on shared benchmarks, and artifacts that can be rerun under a frozen protocol. Many computational sciences provide both, and the Atlas already contains 199 papers whose primary domain is scientific computing and medicine. Experimental sciences will need a different evaluator, such as a validated simulator or a held-out experimental dataset, while the chains, the task packages and the result labels can stay as they are. The cost tasks of the pilot suggest where such work can start: wherever a simulation code or an analysis pipeline has a reference output, a faster or leaner implementation can be checked against it to a stated tolerance.

\section{Conclusion}\label{sec:conclusion}

This report asked whether the published literature can supply tasks for verified research steps in quantity, whether current agents can take such steps, and what confirming them requires. The citation record supplies candidates in quantity: the Paper Atlas places 4,959 of 7,317 papers in 630 task--dataset clusters containing 1,236 citation chains, and a chain segment becomes a task once a builder records the limitation, freezes the protocol and hides the later paper as a reference. Current agents can take some of these steps, and which steps they take is the more useful answer. In the pilot, 16 of 18 submissions on tasks that ask for a cheaper method improved, compared with 8 of 18 on tasks that ask for a better method, and the better methods came mainly where the reference showed a large step. Confirming a step takes more than a better metric. An evaluator has to run every submission, which showed, for example, that one fused kernel never ran; the record has to say whether the agent's own controls support its explanation and whether its route is new; and the step has to be measured under the protocol in which the loop runs, because a published step does not always survive a change of protocol. The next version will draw its tasks from the Atlas, settle every value and rule before the agents run, repeat each run, and let a verified step become the next starting point.

\FloatBarrier

\bibliographystyle{plainnat}
\bibliography{refs}

\appendix
\crefalias{section}{appendix}
\crefalias{subsection}{appendix}
\crefname{appendix}{Appendix}{Appendices}
\Crefname{appendix}{Appendix}{Appendices}
\FloatBarrier
\section{The six pilot chains}\label{app:chains}

This appendix describes each chain of the pilot: the source chain, the setup under which the agents worked and were evaluated, the tasks with their starting points and references, the results with the quantities behind each score, and the failures recorded during evaluation. All values are read from the delivery packages listed in \cref{app:data}. \Cref{tab:design} summarizes how each task implements the design of \cref{sec:tasks}.

\begin{table}[htbp]
  \centering
  \small
  \caption{How each pilot task implements the design. Mode: R research, C claim verification. Reference: a later paper on the chain, the starting method as run in the reproduction package, a model or implementation prepared for the task, or none. Execution: whether the agents could run jobs or wrote code without running it (\cref{sec:pilot-setup}). Scored runs: the seeds or folds behind each value in \cref{tab:results}.}
  \label{tab:design}
  \footnotesize
  \setlength{\tabcolsep}{4pt}
  \begin{tabular}{@{}llllll@{}}
    \toprule
    Chain & Task & Mode & Target & Reference & Execution, scored runs \\
    \midrule
    Density estimation & Efficiency & C & cost & prepared (affine coupling) & jobs, 3 seeds \\
     & Capacity & R & quality & later paper (NAF) & jobs, 3 seeds \\
     & Compactness & C & cost & later paper (B-NAF variant) & jobs, 3 seeds \\
     & Inversion & C & cost & later paper (NSF) & jobs, 3 seeds \\
     & Coupling & R & quality & later paper (LRS) & jobs, 3 seeds \\
    \midrule
    Graph learning & Depth & R & quality & later paper (DropEdge) & jobs, 3 seeds \\
     & Memory & R & cost & prepared (analytic gradients) & jobs, 3 seeds \\
     & Inference & R & cost & prepared (CUDA graph) & jobs, 2 seeds \\
    \midrule
    Action segmentation & A: boundaries & R & quality & later paper (ASRF) & code only, 5 folds \\
     & B: long-range context & R & quality & later paper (ASFormer) & code only, 5 folds \\
     & C: conditioning & R & quality & later paper (ActFusion) & code only, 5 folds \\
    \midrule
    LLM compression & Wanda & R & quality & later paper (DSnoT) & code only, 1 run \\
     & OmniQuant & R & quality & starting method (same value) & code only, 1 run \\
     & SVD-LLM & R & quality & starting method (same value) & code only, 1 run \\
    \midrule
    Semi-supervised & Mean Teacher, 40 labels & R & quality & later paper (FixMatch) & code only, 1 run \\
     & Label propagation & R & quality & full original method & code only, 1 run \\
     & Graph construction & R & cost & starting code (same value) & code only, 3 seeds \\
    \midrule
    Vision backbones & Swin & R & cost & none & code only, 1 run \\
     & Longformer & R & cost & none (timed control) & code only, 1 run \\
     & RepViT & R & cost & none (timed control) & code only, 1 run \\
    \bottomrule
  \end{tabular}
\end{table}

\subsection{Density estimation on MiniBooNE}\label{app:density}

\paragraph{Chain.} MAF \citep{papamakarios2017masked} $\rightarrow$ TAN \citep{oliva2018transformation} $\rightarrow$ NAF \citep{huang2018neural} $\rightarrow$ B-NAF \citep{decao2020block} $\rightarrow$ NSF \citep{durkan2019neural} $\rightarrow$ LRS \citep{dolatabadi2020invertible}, evaluated on the preprocessed 43-dimensional MiniBooNE data used by MAF; TAN did not yield a task. A held-out set of 2,048 rows, drawn from the original training split with a fixed seed, is used only by the evaluator.

\paragraph{Setup.} The tasks ran on one RTX 4090, one job at a time, with each worker limited to a tenth of the GPU memory and two CPU threads, PyTorch 2.7.0, Landlock confinement and no network. The agents worked in stages and launched training jobs through the task's research harness, within per-task limits of 28 jobs, 12,000 optimizer steps, 1,200~s of compute (1,297~s for the inversion task), 12 million parameters and 300 model calls. Every scored run trains for 1,000 steps on seeds 11, 29 and 47.

\paragraph{Score.} The frozen rubric (version 2.1) has seven parts: executability (20 points), baseline (10), claim verification (15), held-out effect (20), cost including NLL non-inferiority (15), mechanism (10) and report consistency (10). The effect part is $20 \cdot \mathrm{clip}(\mathrm{gain}/\mathrm{scale}, 0, 1)$, where scale is the effect scale set in the task and the gain is the drop in held-out NLL for the research tasks and, for the claim-verification tasks, the fractional reduction in inverse time at batch size 64 (efficiency, inversion) or in parameters (compactness); the gain is set to zero if NLL worsens by more than a calibrated margin. Version 2.2, defined after the runs, adds the fraction of the gap to the LRS reference that a submission closes:
\begin{equation*}
  S_{2.2} = 0.7\, S_{2.1} + 30 \cdot \mathrm{clip}\!\left(\frac{v_{\text{start}} - v_{\text{sub}}}{v_{\text{start}} - v_{\text{LRS}}}, 0, 1\right),
\end{equation*}
where $v_{\text{start}}$, $v_{\text{sub}}$ and $v_{\text{LRS}} = 17.442$ are the held-out NLL of the starting point, the submission and LRS. The gap is negative for compactness and inversion, so the version 2.2 report averages only the other three tasks.

\paragraph{Tasks.}
\begin{itemize}
  \item \textbf{Efficiency} (claim verification): reduce sampling latency at batch sizes 1 and 64 and count conditioner calls. Start: an ordered affine autoregressive flow with 16,812 parameters, held-out NLL 20.771 and 21.1~ms for the inverse at batch 64. Reference: an affine coupling model, NLL 25.297.
  \item \textbf{Capacity} (research): test whether a transform more expressive than affine improves likelihood. Start: as above. Reference: NAF, with 46,576 parameters and NLL 18.526.
  \item \textbf{Compactness} (claim verification): verify that the cost of a dense emission grows with the square of the width, then shrink the model. Start: a dense NAF-style conditioner with 547,948 parameters and NLL 16.639. Reference: a B-NAF variant with 30,445 parameters and NLL 24.207, outside the task's likelihood margin.
  \item \textbf{Inversion} (claim verification): reduce the cost of the iterative inverse, with round-trip checks in single and double precision. Start: the same dense model, 2.711~s per inverse. Reference: NSF, NLL 18.299.
  \item \textbf{Coupling} (research): improve the fit of an affine coupling model while keeping inversion cheap. Start: affine coupling with NLL 25.297. Reference: LRS, with 113,686 parameters and NLL 17.442.
\end{itemize}

\begin{table}[htbp]
  \centering
  \small
  \caption{MiniBooNE tasks: the quantity each task targets, held-out NLL (mean over three seeds; lower is better), and scores under the frozen rubric (version 2.1) and the retrospective gap score (version 2.2, three tasks). Scores are given as \gpt{} / \opus.}
  \label{tab:density}
  \begin{tabular}{@{}llrrrrr@{}}
    \toprule
    Task & Quantity & Start & \gpt & \opus & $S_{2.1}$ & $S_{2.2}$ \\
    \midrule
    Efficiency & inverse at batch 64 (ms) & 21.1 & 16.27 & 3.56 & 84.69 / 97.03 & 59.28 / 78.16 \\
               & NLL & 20.771 & 20.771 & 19.635 & & \\
    Capacity & NLL & 20.771 & 16.111 & 16.681 & 100 / 100 & 100 / 100 \\
    Compactness & parameters & 547,948 & 279,628 & 106,604 & 90.37 / 75.00 & --- \\
                & NLL & 16.639 & 16.524 & 18.315 & & \\
    Inversion & inverse time (s) & 2.711 & 1.031 & 0.217 & 92.45 / 98.49 & --- \\
              & NLL & 16.639 & 16.639 & 15.061 & & \\
    Coupling & NLL & 25.297 & 25.163 & 18.499 & 80.35 / 97.85 & 56.76 / 94.46 \\
    \midrule
    \multicolumn{5}{@{}l}{Mean (version 2.1: five tasks; version 2.2: three tasks)} & 89.57 / 93.67 & 72.01 / 90.87 \\
    \bottomrule
  \end{tabular}\\[3pt]
  \parbox{\linewidth}{\footnotesize\raggedright The \opus{} compactness model fails the likelihood margin, so its effect part is zero.}
\end{table}

\paragraph{What the agents did.} On efficiency, \opus{} shared autoregressive degrees across blocks of coordinates, so that its inverse needs 16 sequential conditioner calls instead of the 172 of the same model without blocks; its ablation confirmed that this caused the speed-up. \gpt{} left the density unchanged, optimized only the inverse, and rejected its own claim after its ablation turned out faster. On capacity, \opus{} used a sum of inverse hyperbolic sine functions and \gpt{} a bounded tanh residual; both reached a lower NLL than the reference, NAF, and than LRS, with 150,972 and 83,892 parameters respectively, within the limit of 12 million. On compactness, \gpt{} used a rank-one emission whose cost grows linearly with the width, and \opus{} used a rank-16 emission that failed the likelihood margin, which it reported. On inversion, \opus{} used an autoregressive spline with an analytic inverse, and \gpt{} applied Newton's method to the unchanged model; the speed-up it had registered for Newton's method over its own bisection ablation did not appear at batch 64. On coupling, the 1,109,740-parameter \opus{} model did worse than its own affine ablation, which refuted its spline claim, and the warp used by \gpt{} missed its own threshold.

\paragraph{Evaluation records.} All ten results are unassisted, and each passed a separate re-evaluation and a fresh-training replay. The recorded failures were infrastructure faults, each retried once: CUDA blocked by an outer sandbox in the first \opus{} efficiency attempt, which restarted from scratch; two launch failures of the Codex client, one from an OpenSSL permission error and one from a missing home-directory setting; and bursts of server errors from the model provider, 11 consecutive failed calls in each of two \opus{} tasks. A ruling that set the \opus{} efficiency score to zero for departing from its registered plan was reversed, because the rule had been introduced after the runs. Stage wall-clock time including retries was 11,803~s for \opus{} and 6,056~s for \gpt{}, with 452 and 289 model calls.

\subsection{Graph learning on Cora}\label{app:cora}

\paragraph{Chain.} GCN \citep{kipf2017semisupervised} $\rightarrow$ GAT \citep{velickovic2018graph} $\rightarrow$ DropEdge \citep{rong2020dropedge} $\rightarrow$ GRAND \citep{feng2020graph} $\rightarrow$ NAGphormer \citep{chen2023nagphormer} $\rightarrow$ Classic GNNs \citep{luo2024classic}, all evaluated on node classification on Cora \citep{sen2008collective}, with the author repositories of all six papers pinned to commits. The data are the full citation graph, with 2,708 nodes, 1,433 features, 7 classes and 5,278 undirected edges, under the standard split \citep{yang2016revisiting} of 140 training, 500 validation and 1,000 test nodes; the test labels are private.

\paragraph{Setup.} The tasks ran on one RTX 4090 with at most two CPU cores and no network, each worker confined by Landlock and seccomp under a separate user account. As in the density chain, and unlike the four chains that follow, the agents could launch a small number of jobs through the task's research harness: two complete training jobs for the depth task, two submission rounds for the memory task and two public validation runs for the inference task. The evaluator then retrained or re-ran each sealed submission from scratch.

\paragraph{Score.} The three tasks share one rule. For seed $i$, let $B_i$, $R_i$ and $P_i$ be the values of the starting point, the reference and the submission on a metric where lower is better, with $\mathrm{gain}_i = B_i - P_i$ and $\mathrm{gap}_i = B_i - R_i$. Over $k$ seeds,
\begin{equation}
  S = 2.5\,n + \frac{10}{k}\sum_{i=1}^{k} \mathbf{1}\!\left[\mathrm{gain}_i > g\right] + \frac{30}{k}\sum_{i=1}^{k} \mathrm{clip}\!\left(\frac{\mathrm{gain}_i}{\mathrm{gap}_i}, 0, 1\right),
  \label{eq:cora}
\end{equation}
where $\mathbf{1}[\cdot]$ is the indicator function, $n \le 24$ counts the evidence items met, and the threshold $g$ is 0.02~nats, 16~MiB and 1~\textmu s for the depth, memory and inference tasks, respectively. The 24 evidence items form six groups of four, covering reproduction, the baseline, verification of the problem, cost, mechanism and the consistency of the report. They were judged by a separate instance of Codex running \gpt.

\paragraph{Tasks.} All three are research tasks.
\begin{itemize}
  \item \textbf{Depth degradation.} The starting point is nine trained GCNs of depth 2, 8 and 16, three seeds each. The agent registers a falsifiable intervention before running it, then trains the proposal, an ablation, a perturbation and matched controls for 200 updates each. The metric is the test cross-entropy averaged over depths 8 and 16. The reference is DropEdge, continued from the same checkpoints with 40\% of the edges kept in each epoch.
  \item \textbf{Training memory.} The starting point is an exact CPU workload: a two-layer sparse GCN trained from three checkpoints for 200 full-batch steps with validation at each step, a checkpoint reload and final logits. The submission must lower peak resident memory while matching the starting trajectory to $10^{-5}$ for 200 steps and staying within twice its runtime. The reference is an analytic-gradient implementation in NumPy and SciPy that recomputes intermediate values.
  \item \textbf{Inference scheduling.} The starting point is a fixed, untrained eight-head GAT with 92,391 parameters. The submission may change only scheduling and buffer ownership, must keep every logit within $10^{-5}(1 + |\ell|)$ of the starting logit $\ell$ and produce exactly the same predictions, and may not cache results. The metric is the synchronized latency of a request. The reference is a single CUDA graph capture of the unchanged model.
\end{itemize}

\begin{table}[htbp]
  \centering
  \small
  \caption{Cora tasks: seed means of each metric (lower is better) and scores (\cref{eq:cora}).}
  \label{tab:cora}
  \begin{tabular}{@{}lrrrrrr@{}}
    \toprule
    Task (metric) & Start & Reference & \gpt & \opus & $S$ (\gpt) & $S$ (\opus) \\
    \midrule
    Depth (cross-entropy, nats) & 1.620 & 1.562 & 1.640 & 1.661$^{s}$ & 58.00 & 53.96$^{s}$ \\
    Memory (peak memory, MiB) & 484.7 & 53.2 & 422.0 & 56.1$^{a,s}$ & 74.35 & 94.80$^{a,s}$ \\
    Inference (latency, \textmu s) & 656.3 & 382.0 & 391.6 & 385.8$^{a}$ & 96.45 & 94.55$^{a}$ \\
    \bottomrule
  \end{tabular}\\[3pt]
  \parbox{\linewidth}{\footnotesize\raggedright $^{s}$ second attempt from the unchanged public package, after the first was rejected at the method review (depth) or failed in the agent's own code (memory). $^{a}$ assisted result; the unassisted results of \opus{} were a failed run (memory) and a score of 0 under the task's rule (inference).}
\end{table}

\paragraph{What the agents did.} On the depth task, \gpt{} mixed the GCN propagation operator with the identity, and \opus{} used a lazy propagation operator $(1-\alpha)P + \alpha I$, where $P$ is the GCN propagation matrix and $\alpha$ a mixing weight. Both lowered the cross-entropy for the first seed only, and neither claimed that its mechanism held across seeds; the evaluator's fresh retraining reproduced both metrics within $5 \times 10^{-7}$. On the memory task, \gpt{} replaced the PyTorch optimizer with an explicit single-tensor Adam, which lowered peak memory by 13\%, and \opus{} replaced PyTorch with a NumPy and SciPy engine, the approach of the hidden reference. On the inference task, \gpt{} replayed the forward pass as a CUDA graph and \opus{} combined a fused attention layer with graph replay; both came within 3\% of the reference latency on average over the two seeds.

\paragraph{Evaluation records.} The first \opus{} depth submission was rejected at the evaluator's method review: its validation had selected the arm with $\alpha = 0.45$, but the predictions it submitted ran $\alpha = 0.15$, so the scored model differed from the one its diagnostics described. The reported result comes from a second attempt that started again from the unchanged public package, without feedback from the first; its first training run failed on a gradient-diagnostic error and counted against its budget. On the memory task, the first \opus{} run failed because its own code set the PyTorch thread count twice. A second attempt from the unchanged public package then failed the trajectory check at one step, with an error of $1.05 \times 10^{-5}$ where the limit is $10^{-5}$; the assisted result follows a repair of that attempt by the operators, who replaced one sparse accumulation routine and corrected the report's attribution of the ablation. On the inference task, the unassisted \opus{} submission failed the pre-execution audit twice because of errors in the agent's setup code, among them a device comparison that rejected valid inputs and a graph capture outside the 30~s initialization limit; a failed audit scores zero under the task's rule. The assisted result follows a repair of the device check, of the capture placement and of the preparation entry point, with the arithmetic untouched. The \gpt{} inference run first failed the same audit because the operators had omitted construction evidence and one registered public replay; after they supplied both, with the submitted source unchanged, the run passed.

\subsection{Temporal action segmentation on 50 Salads}\label{app:tas}

\paragraph{Chain.} MS-TCN \citep{abufarha2019mstcn} $\rightarrow$ ASRF \citep{ishikawa2021alleviating} $\rightarrow$ ASFormer \citep{yi2021asformer} $\rightarrow$ UVAST \citep{behrmann2022unified} $\rightarrow$ DiffAct \citep{liu2023diffusion} $\rightarrow$ ActFusion \citep{gong2024actfusion}, evaluated on 50 Salads \citep{stein2013combining} with pre-extracted I3D features \citep{carreira2017quo}, 19 classes and five folds. The six papers form a graph with 15 citation edges, and the chain is its longest path. The reproduction package trained MS-TCN, ASRF, ASFormer and DiffAct on all five folds and reached F1@50 scores of 64.58, 74.67, 79.48 and 83.81.

\paragraph{Setup.} As on the other code-only chains (\cref{sec:pilot-setup}), the agents wrote code without executing it, here with a separate budget for each fold: 64 model calls and 14,400~s. The code for each fold was sealed before any training. The evaluator then trained and tested each fold with one GPU, eight CPU cores and 16~GiB of memory, within 7,200~s (2~h) for the first task and 43,200~s (12~h) for the other two. The metric is F1@50, the segment-level F1 score at 50\% overlap, computed with the original evaluation code on the ten test videos of each fold and averaged over the five folds.

\paragraph{Tasks.} All three are research tasks.
\begin{itemize}
  \item \textbf{A, boundaries in MS-TCN.} The starting point is the MS-TCN code trained for 50 epochs; the limitation is over-segmentation, that is, short false segments and misplaced boundaries. The reference is ASRF (74.67). The package's value for the starting point, 64.58, was measured on all frames, whereas this task scores every second frame.
  \item \textbf{B, long-range context in ASRF.} The starting point is ASRF trained from scratch for 80 epochs on a 36/4/10 split into training, validation and test videos, with the checkpoint chosen by validation edit score; the limitation is that local convolutions confuse distant actions. The reference is ASFormer (79.48). The package's value for the starting point, 74.67, comes from a different training protocol.
  \item \textbf{C, conditioning in DiffAct.} The starting point is DiffAct trained from scratch for 5,001 epochs with 25 sampling steps; the limitation is that weak conditioning at some positions hurts consistency across the whole video. The starting point scores 83.81 and the reference, the authors' ActFusion checkpoints, 84.81.
\end{itemize}

\begin{table}[htbp]
  \centering
  \small
  \caption{Temporal action segmentation on 50 Salads: F1@50 (\%, higher is better), mean over five folds.}
  \label{tab:tas}
  \begin{tabular}{@{}lrrrr@{}}
    \toprule
    Task & Start & Reference & \gpt & \opus \\
    \midrule
    A, boundaries in MS-TCN & 64.58$^{*}$ & 74.67 & 60.04 & 60.22 \\
    B, long-range context in ASRF & 74.67$^{*}$ & 79.48 & 73.88 & 72.32$^{r}$ \\
    C, conditioning in DiffAct & 83.81 & 84.81 & 84.37 & 82.40$^{r}$ \\
    \bottomrule
  \end{tabular}\\[3pt]
  \parbox{\linewidth}{\footnotesize\raggedright $^{*}$ measured under a different protocol from the task's; $^{r}$ recovered result, completed in an additional engineering window.}
\end{table}

\paragraph{What the agents did.} On task A, \gpt{} left the network unchanged and modified only the loss terms around boundaries, while \opus{} added a trained boundary branch with boundary-gated refinement; the first fold of \opus{} scored only 44.18. On task B, \gpt{} added gated attention over a pooled summary of the whole video, and \opus{} added zero-initialized gated self-attention over pooled time steps. On task C, \gpt{} added gated multi-scale conditioning signals after DiffAct's masks, and \opus{} trained with contiguous masks of the conditioning, explicit mask indicators and a summary of the whole video.

\paragraph{Evaluation records.} A CPU rescoring of the saved predictions of all 30 fold runs reproduced every fold metric and all six scores exactly. On task B, after the agent had revised the code of two \opus{} folds, the operators failed to launch them before their deadlines, and one window of 3,600~s per fold was authorized; on task C, a GPU-monitoring sampler stopped one \opus{} fold three times, and the fold finished in a fresh 43,200~s window. Both agents also fixed errors in their own code for individual folds, such as an environment variable read without being set and a prediction file missing a required field; such fixes count against the agents' budgets.

\subsection{Language-model compression on WikiText-2}\label{app:llm}

\paragraph{Chain.} SmoothQuant \citep{xiao2023smoothquant} $\rightarrow$ SparseGPT \citep{frantar2023sparsegpt} $\rightarrow$ LLM-Pruner \citep{ma2023llmpruner} $\rightarrow$ Wanda \citep{sun2024simple} $\rightarrow$ DSnoT \citep{zhang2024dynamic} $\rightarrow$ OmniQuant \citep{shao2024omniquant} $\rightarrow$ PB-LLM \citep{yuan2024pbllm} $\rightarrow$ BiLLM \citep{huang2024billm} $\rightarrow$ SVD-LLM \citep{wang2025svdllm}, with a branch DSnoT $\rightarrow$ SLEB \citep{song2024sleb}, all evaluated on WikiText-2 \citep{merity2017pointer} with LLaMA-7B \citep{touvron2023llama}. The chain is maximal in a graph of 27 WikiText-2 papers and 96 direct citation edges; the graph's longest path has 13 papers. The chain's reproduction package ran all ten methods within 48~h, and seven matched the perplexity reported in their papers within 5\%.

\paragraph{Setup.} The tasks ran on one RTX 4090 with 24~GB, PyTorch 2.2.0 and CUDA 12.1, with LLaMA-7B in FP16. The agents wrote code without executing it. The evaluation windows were 7,200~s (2~h) for Wanda and 14,400~s (4~h) for OmniQuant and SVD-LLM. The metric is WikiText-2 test perplexity over 166 non-overlapping blocks of 2,048 tokens. The scorer rebuilds the compressed model from the submitted numbers alone, that is, masks, clipping bounds or factors, and never imports the submitted code.

\paragraph{Tasks.} All three tasks are research tasks; their reference values come from the reproduction package and are used for comparison only.
\begin{itemize}
  \item \textbf{Wanda at 60\% sparsity.} The starting point is Wanda's one-shot importance ranking, whose limitation is that it does not minimize the output error of the pruned layer. The submission supplies masks for all 224 projections with a fixed number of pruned weights per row and may not update weights. The hidden reference is DSnoT.
  \item \textbf{OmniQuant with 3-bit weights.} The starting point is OmniQuant's learnable clipping, trained for 20 epochs, whose limitation is that uniform initialization and a plain block-wise squared error may miss good clipping bounds at 3 bits. The submission supplies the 448 clipping tensors. The reference is the package's OmniQuant result.
  \item \textbf{SVD-LLM whitening at 20\% compression.} The starting point is SVD-LLM's whitened truncation with fixed ranks, whose limitation is that a locally weighted fit ignores how errors propagate across layers. The submission supplies 448 factors at the fixed ranks, without recovery fine-tuning. The reference is the package's whitening-stage result.
\end{itemize}

\begin{table}[htbp]
  \centering
  \small
  \caption{Language-model compression: WikiText-2 test perplexity (lower is better). For OmniQuant and SVD-LLM the reference is the starting method.}
  \label{tab:llm}
  \begin{tabular}{@{}lrrrr@{}}
    \toprule
    Task & Start & Reference & \gpt & \opus \\
    \midrule
    Wanda, 60\% sparsity & 10.705 & 10.678 & 11.737 & 12.416 \\
    OmniQuant, W3A16 & 6.503 & 6.503 & 6.514 & 6.595$^{r}$ \\
    SVD-LLM whitening & 7.892 & 7.892 & 7.890 & 7.424 \\
    \bottomrule
  \end{tabular}\\[3pt]
  \parbox{\linewidth}{\footnotesize\raggedright $^{r}$ recovered result, completed in an additional engineering window.}
\end{table}

\paragraph{What the agents did.} On Wanda, both agents started from the one-shot ranking and swapped kept and pruned weights to lower an estimate of the layer's output error. \gpt{} used an inexpensive diagonal estimate, and perplexity rose by 1.03. \opus{} used the exact error computed from the full input correlation matrix and cut that per-layer error by 20.7\% across all projections, yet perplexity rose by 1.71; its own falsification test, stated before the run, had named this outcome: the per-layer error does not predict perplexity. On OmniQuant, \gpt{} searched 25 pairs of initial clipping bounds per channel before the original optimization, and \opus{} used a two-sided clipping search weighted by input energy with a normalized block loss; both raised perplexity, to 6.514 and 6.595 from 6.503. On SVD-LLM, \gpt{} regularized the activation covariance before whitening and left perplexity essentially unchanged (7.890 compared with 7.892), while \opus{} fitted each factor to the activations of the already compressed model and let the output projections correct the drift from the dense model, lowering perplexity by 0.47.

\paragraph{Evaluation records.} The \gpt{} Wanda run met three runner failures (an unimportable starter module, CUDA blocked by a nested sandbox, and a permission error) and one error in its own code, which the agent fixed in a revision after seeing only its error trace; the final run finished within the original deadline. The original \opus{} OmniQuant run was stopped by infrastructure faults, including a monitor that terminated a healthy process, after the agent had added a missing function to its own code; the operators authorized one additional window of 10,800~s (3~h), in which the run finished.

\subsection{Semi-supervised learning on CIFAR-10}\label{app:ssl}

\paragraph{Chain.} Mean Teacher \citep{tarvainen2017mean} $\rightarrow$ Label Propagation \citep{iscen2019label} $\rightarrow$ Noisy Student \citep{xie2020selftraining} $\rightarrow$ FixMatch \citep{sohn2020fixmatch} $\rightarrow$ DARP \citep{kim2020distribution} $\rightarrow$ CReST \citep{wei2021crest} $\rightarrow$ CoSSL \citep{fan2022cossl}: seven papers linked by twelve citation edges, with the chain as the graph's longest path. No single dataset covers all seven papers, and the runs use CIFAR-10 \citep{krizhevsky2009learning}. The reproduction package executed six of the seven papers; Label Propagation reached 88.38\% and FixMatch 92.99\%.

\paragraph{Setup.} The agents wrote code without executing it, with 64 model calls and 14,400~s per task. The evaluator ran each sealed submission on an RTX 4090 within 129,600~s (36~h) for the two training tasks and 3,600~s (1~h) for the graph task. Following the original code, a classification result is the best test accuracy among the evaluations made during training, and the selected state is replayed; this rule applies to the starting Mean Teacher run, the references and the submissions. For label propagation, the starting value is the accuracy at the end of the first stage, from which the second stage begins.

\paragraph{Tasks.} All three are research tasks.
\begin{itemize}
  \item \textbf{Mean Teacher with 40 labels.} The starting point is Mean Teacher with a WRN-28-2 network trained from random initialization for 262,144 steps on the 40-label split; with four labels per class, its predictions on unlabeled images are unreliable. The submission may change the loss, the handling of confidence, the use of the teacher and of pseudo-labels, and the augmentation. The starting point reaches 23.88\% and the reference, FixMatch, 92.99\%.
  \item \textbf{Label propagation with 4,000 labels.} The starting point trains the original network for 180 epochs on the labeled images; a second stage of 180 epochs must make use of the other 46,000 images. The submission supplies the second-stage method and may change only pseudo-targets and sample or class weights. The first stage ends at 79.37\% and 79.02\% in the two agents' runs; the reference, the original method, reaches 88.38\%.
  \item \textbf{Label-propagation graph construction.} The starting point builds the exact $k$-nearest-neighbor graph used by Label Propagation for 50,000 vectors of dimension 128 with $k = 50$ on the CPU \citep{johnson2021billion}, in 9.5 to 9.8~s per call. The submission must reproduce the graph exactly, without approximate search, and is scored by \cref{eq:timesaving} (\cref{app:arch}) over three seeds; the starting code is also the reference.
\end{itemize}

\begin{table}[htbp]
  \centering
  \small
  \caption{Semi-supervised learning on CIFAR-10: top-1 accuracy (\%) and, for graph construction, the speed-up $G$ over the starting code (higher is better).}
  \label{tab:ssl}
  \begin{tabular}{@{}lrrrr@{}}
    \toprule
    Task & Start & Reference & \gpt & \opus \\
    \midrule
    Mean Teacher, 40 labels & 23.88 & 92.99 & 34.23 & 76.34 \\
    Label propagation, 4,000 labels & 79.37 / 79.02$^{\dagger}$ & 88.38 & 87.30 & 88.61 \\
    Graph construction ($G$) & 1 & 1 & 8.26 & 95.48 \\
    \bottomrule
  \end{tabular}\\[3pt]
  \parbox{\linewidth}{\footnotesize\raggedright $^{\dagger}$ first-stage accuracy at the end of the stage in the \gpt{} and \opus{} runs.}
\end{table}

\paragraph{What the agents did.} With 40 labels, \gpt{} combined hard pseudo-labels on strongly augmented images above a confidence of 0.95 with a weighted consistency loss below it, using seven times as many unlabeled images per step. \opus{} used three terms: a teacher-to-student pseudo-label loss under an adaptive soft confidence gate, class-distribution alignment and a negative-label loss. For label propagation, \gpt{} built pseudo-labels from running averages of predictions filtered by confidence and agreement, without a graph, while \opus{} propagated labels over a nearest-neighbor graph of the current embeddings in every epoch and went past the reference. For graph construction, \gpt{} ran a blocked exhaustive search in double precision on the GPU and built the sparse graph on the CPU, and \opus{} ran a blocked single-precision search with exact top-$k$ selection and built the graph on the GPU; both passed the exactness checks on all three seeds.

\paragraph{Evaluation records.} The \gpt{} run with 40 labels took 113,070~s because its augmentation ran serially on the CPU. The first \opus{} 40-label run stopped at step 1,024 because the operators' evaluation request pointed to the wrong machine, and it was rerun from scratch. Replays of the selected states reproduced the accuracies of both label-propagation runs and of the \gpt{} run with 40 labels.

\subsection{Vision backbones: three operator tasks}\label{app:arch}

\paragraph{Chain.} ViT \citep{dosovitskiy2021image} $\rightarrow$ T2T-ViT \citep{yuan2021tokens} $\rightarrow$ PVT \citep{wang2021pyramid} $\rightarrow$ Swin \citep{liu2021swin} $\rightarrow$ Vision Longformer \citep{zhang2021multiscale} $\rightarrow$ NesT \citep{zhang2022nested} $\rightarrow$ RegionViT \citep{chen2022regionvit} $\rightarrow$ CrossFormer \citep{wang2022crossformer} $\rightarrow$ EfficientFormer \citep{li2022efficientformer} $\rightarrow$ SwiftFormer \citep{shaker2023swiftformer} $\rightarrow$ RepViT \citep{wang2024repvit}, all evaluated on ImageNet-1K \citep{deng2009imagenet}. The chain is the unique longest path of an eleven-paper subgraph with 43 citation edges, selected from a candidate graph of 90 ImageNet-1K classification papers with 453 edges and 27 tied longest paths. The three tasks study operators of three papers on the chain and involve no ImageNet training; they reuse the pinned author code of the environment package described in \cref{sec:startpoint} but were run in a separate runtime.

\paragraph{Setup.} The tasks ran on one RTX 4090 with PyTorch 2.5.1 and CUDA 12.4. The agents wrote code without executing it (\cref{sec:pilot-setup}). The evaluator ran each sealed submission within a single 3,600~s window per submission that covered compilation, correctness checks and timing.

\paragraph{Score.} With $G$ the geometric mean, over the test conditions, of the runtime ratio $T_{\text{start}}/T_{\text{submission}}$, the time-saving score is
\begin{equation}
  S = 100 \cdot \max\!\left(0,\; 1 - 1/G\right).
  \label{eq:timesaving}
\end{equation}
Each runtime is the median of five alternating repetitions after 20 warm-up iterations, and a correctness failure scores zero. The starting operator itself has $G = 1$ and scores zero.

\paragraph{Tasks.}
\begin{itemize}
  \item \textbf{Swin window shift and partition.} The starting point is the unfused PyTorch path: a cyclic shift, a window partition, and the reverse shift. The limitation is that the shift and the permutation are separate operations that materialize copies. The submission must produce outputs and input gradients identical to the starting point, in FP32 and FP16 and in both directions, at batch 192, resolution 56$\times$56, 96 channels, shift 2 and window 7. The authors' unit test is part of the private checks. The authors' fused kernel was part of the public package as study material and was not timed; the task has no reference value.
  \item \textbf{Vision Longformer local attention.} The starting point is the original sliding-chunk attention with automatic differentiation, which stores nine copies of the features for the backward pass. The submission must reproduce the same attention, checked against masked full attention on 100 random inputs, and run the forward and backward passes faster. The authors' version with a hand-written backward pass was part of the public package and is timed as a control, not scored.
  \item \textbf{RepViT depthwise reparameterization.} The starting point is the unfused depthwise block in inference mode: a batch-normalized 3$\times$3 convolution, a 1$\times$1 convolution and an identity branch. The submission must fold the block into one operator, with a derivation of the fold, and match the starting point to $10^{-4}$ on nine batch-norm states. The authors' fusion function is timed as a control.
\end{itemize}

\begin{table}[htbp]
  \centering
  \small
  \caption{Operator tasks: geometric-mean speed-up $G$ and time-saving score $S$ (\cref{eq:timesaving}). The last column gives the runtime of the starting operator and the submission, or the comparison with the authors' timed control.}
  \label{tab:arch}
  \begin{tabular}{@{}llrrl@{}}
    \toprule
    Task & Agent & $G$ & $S$ & Runtime or comparison with the control \\
    \midrule
    Swin & \gpt & 3.087 & 67.61 & 3.01 $\rightarrow$ 1.01~ms (FP32 partition) \\
         & \opus & 1.000 & 0.00 & fell back to the starting path \\
    Vision Longformer & \gpt & 1.693 & 40.94 & 8.08 $\rightarrow$ 4.77~ms \\
         & \opus & 2.774 & 63.95 & 6.62 $\rightarrow$ 2.39~ms \\
    RepViT & \gpt & 6.654 & 84.97 & on par with the authors' fusion \\
         & \opus & 11.270 & 91.13 & 1.68 times as fast as the authors' fusion \\
    \bottomrule
  \end{tabular}
\end{table}

\paragraph{What the agents did.} On Swin, \gpt{} applied the shift and the window partition as one permutation in a vectorized CUDA copy kernel, and its backward pass applied the inverse permutation; all 24 exact checks passed. \opus{} compiled a single-gather CUDA kernel guarded by a self-check, but the check computed gradients inside the forward pass of a custom autograd function, where gradients are disabled, so every check failed and the code fell back to its own copy of the starting path. The result is correct and no faster ($G = 0.99993$). On Vision Longformer, \gpt{} merged the nine neighbor-block matrix products and wrote an explicit backward pass, while \opus{} used one padded key--value buffer with three overlapping strided products, a compiled mask-and-softmax and a backward pass that saves four tensors; both reduced peak allocated memory from about 550~MB to about 403~MB. On RepViT, both agents folded the branches exactly into one depthwise convolution, and \opus{} added a custom 3$\times$3 depthwise CUDA kernel.

\paragraph{Evaluation records.} The two Longformer starting operators were timed in different runs (8.08 and 6.62~ms), and the RepViT measurements used four CPU threads for \gpt{} and one for \opus{}. The first attempt of \opus{} on Longformer reached the 1,800~s limit of a single call with no accepted file; its continuation counted against the same budget. The first timing of the \gpt{} Longformer submission failed in the authors' control rather than in the submission; only the control was fixed, and the submission's first measurement stands. The first launch of the \gpt{} RepViT run failed before any submitted code ran, because a system tool was missing. Five evaluation runs of the \opus{} RepViT submission met build-environment faults (a missing build tool, a refused library link, a blocked process spawn and a missing CUDA header), and the code used its fallback path; after fixes to the environment alone, the sixth run used the submitted kernel. The fallback scores of the earlier runs, 85.9 to 86.6, are kept in the record and not used as the score.

\FloatBarrier
\section{Data and artifacts}\label{app:data}

The Paper Atlas snapshot of 6 November 2026, with its paper records, clusters, display chains and counting rules, can be browsed at \url{https://rsi-atlas.ai}, where the distribution workbook and the summary data can also be downloaded. The pilot's results come from six delivery packages held by the project, one per chain: density estimation (27 October 2026), graph learning (9 November 2026), and action segmentation, language-model compression, semi-supervised learning and the operator tasks (8 November 2026 each). Each package contains the public task statements, the private references and scorer code, the agents' observable trajectories, the evaluator's records and a SHA-256 manifest of every file. Their public parts can be released as task packages, while the private parts, which contain the hidden references, stay with the evaluators.

\section{Contributors}\label{app:contributors}

This report was prepared by the RSI AI4AI Research Team. The contributors and their affiliations are listed below.

\begin{center}
\begin{tabular}{@{}ll@{}}
  \toprule
  Contributor & Affiliation \\
  \midrule
  Jie Xiao & Gradient \\
  Pengbin Feng & University of Southern California \\
  Xinge Liu & University of Toronto \\
  Jingwei Shi & The Hong Kong Polytechnic University \\
  SimonT & Gradient \\
  Jingwei Song & Shanghai Jiao Tong University \\
  Wanyi Chen & Soochow University, Suzhou \\
  Haixin Wang & University of California, Los Angeles \\
  Qingnan Ren & Gradient \\
  Eric Yang & Gradient \\
  Jingsong Yang & Peking University \\
  Huan Zhang & Université de Montréal, Mila – Quebec Artificial Intelligence Institute \\
  Tianyu Shi & McGill University \\
  \bottomrule
\end{tabular}
\end{center}

\end{document}